\documentclass{article}

\usepackage[final]{neurips}

\usepackage[utf8]{inputenc} 
\usepackage[T1]{fontenc}    
\usepackage{hyperref}       
\usepackage{url}            
\usepackage{booktabs}       
\usepackage{amsfonts}       
\usepackage{nicefrac}       
\usepackage{microtype}      
\usepackage{graphicx}
\usepackage{algorithm}
\usepackage{algorithmic}
\usepackage{tabularx}
\usepackage{makecell}
\usepackage{array}
\newcolumntype{Y}{>{\centering\arraybackslash}X}
\usepackage{enumitem}

\usepackage{amsmath}
\usepackage{amssymb}
\usepackage{mathtools}
\usepackage{amsthm}

\title{An Approach to Solving the ARC Challenge}

\author{%
 John Chong Min Tan \\
  Department of Electrical and Computer Engineering\\
  National University of Singapore\\
  \texttt{johntancm@u.nus.edu} \\
}

\begin{document}

\maketitle

\begin{abstract}
We utilise the power of Large Language Models (LLMs), in particular GPT4, to be prompt engineered into performing an arbitrary task. Here, we give the model some human priors via text, along with some typical procedures for solving the ARC tasks, and ask it to generate the i) broad description of the input-output relation, ii) detailed steps of the input-output mapping, iii) use the detailed steps to perform manipulation on the test input and derive the test output. The current GPT3.5/GPT4 prompt solves 2 out of 4 tested small ARC challenges (those with small grids of 8x8 and below). With tweaks to the prompt to make it more specific for the use case, it can solve more. We posit that when scaled to a multi-agent system with usage of past memory and equipped with an image interpretation tool via Visual Question Answering, we may actually be able to solve the majority of the ARC challenge.
\end{abstract}

\section{Background}

The ARC Challenge is a very interesting challenge, as it is doing something counter to mainstream deep learning – learning from very few samples. Deep learning typically uses tens of thousands of samples to do well, for instance learning to classify digits (MNIST) \citep{deng2012mnist} requires around 50,000 training samples. Humans, in comparison, can learn how to identify different animals by just one or two different observations. For instance, my 3 year-old kid can learn how to identify a giraffe in real life for the first time, even though the only other time he was exposed to a giraffe was through a cartoon flash card. Such capabilities are not well endowed in modern AI systems, and that means that such AI systems will need to be trained extensively before deploying in the real world. After deploying them in the real world, they will also be limited in their ability to adapt and learn as the environment changes.

In contrast, traditional rule-based systems (e.g. GOFAI) can “learn” quite fast, as any new situation can be interpreted without any learning phase, provided that the situation is already in the system rules given to it. Such a rule-based system could be symbolic systems or expert systems which already have the domain knowledge fed to it by human experts. However, the history of GOFAI has shown that it is difficult to engineer these rules out, and at many times, even humans face difficulty to come up with the rules as they may not be able to express it in words.

As you can see, there are shortcomings with the above two approaches, and a new kind of approach will need to be used in order to learn fast and generalise to new situations, in order to even have a chance at solving the ARC Challenge.

\section{Next token prediction for self-supervised learning}

\begin{figure}[t]
\centering
		\includegraphics[width=\textwidth]{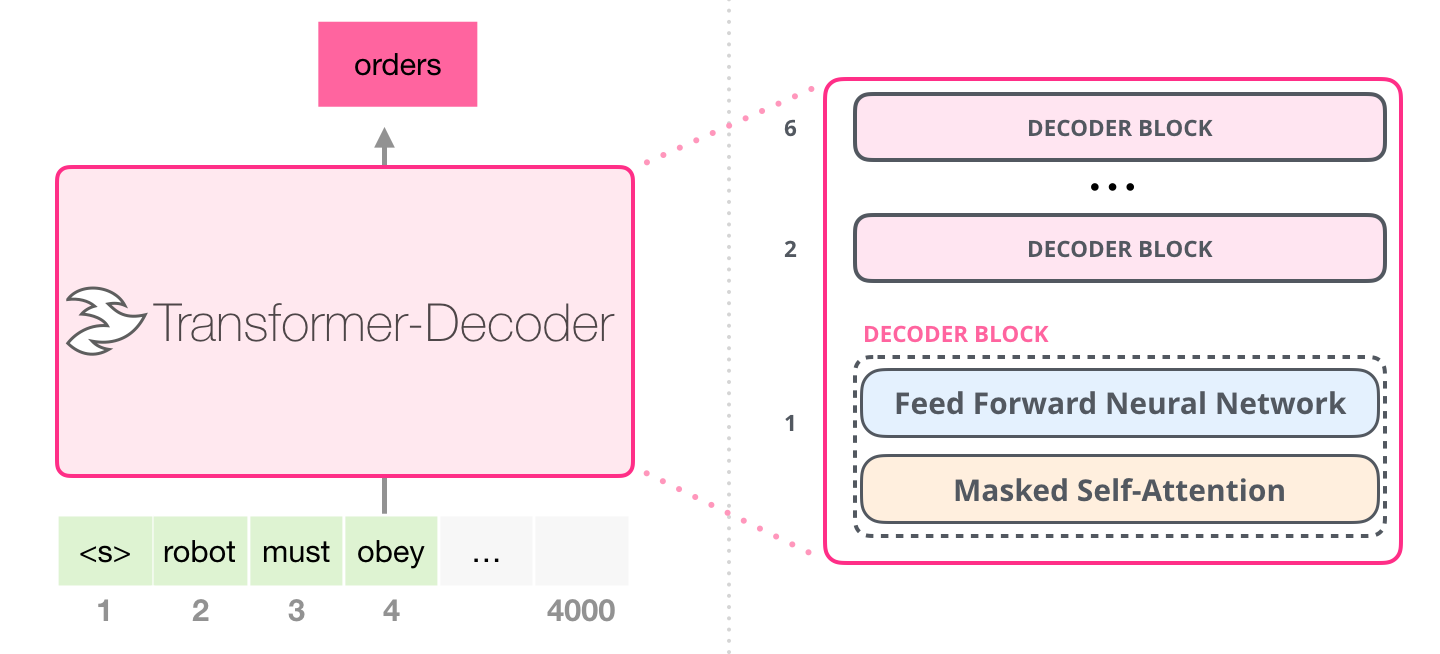}
		\caption{Transformer next-token prediction framework illustration. Here, for simplicity, we just denote the tokens as words, when in actual fact, it should be at the sub-word level built up using a variant of Byte Pair Encoding (BPE) \citep{shibata1999byte}. Image taken taken from: \url{https://jalammar.github.io/illustrated-gpt2/} \citep{jalammar-gpt2}}
		\label{fig:Transformer-GPT}
\end{figure}

There is a lot of structure in the world. These structures can be hard to represent via verbal rules, yet children can learn how physics work and how to interact with the world just by observation and action. Personally, I believe that simply observing is not enough – one has to perform actions in order to learn how one’s actions can affect the world. However, for tasks like learning language, the next action to take is simply to predict the next token and can be done without interaction with the world. Large Language Models (LLMs) such as GPT2 \citep{radford2019language}, GPT 3.5 \citep{ouyang2022training} and GPT4 \citep{bubeck2023sparks} have utilised an extensive amount of self-supervised learning via next-token prediction in order to learn the structure of text (See Fig. \ref{fig:Transformer-GPT}). This is a huge breakthrough, as the predominant approach to deep learning - supervised learning - requires extensive human labelling and is expensive and impractical to obtain for large amounts of data. This self-supervised learning approach can generate labels simply by predicting the next token and is easily obtainable from the world's worth of text on the World Wide Web. For instance, the sentence "The cat sat on the mat" can easily be used in at least 5 different prediction tasks (assuming tokens are defined at the word level), as shown below:

\begin{enumerate}
    \item $The\ \rightarrow cat$
    \item $The\ cat\ \rightarrow sat$
    \item $The\ cat\ sat\ \rightarrow on$
    \item $The\ cat\ sat\ on\ \rightarrow the$
    \item $The\ cat\ sat\ on\ the\ \rightarrow mat$
\end{enumerate}

\textbf{High sample efficiency.} This means that the observations from the world can be reused in multiple input-output pairs and there is very high sample efficiency due to such a self-supervised learning method able to reuse the same sections of text multiple times. 

\begin{figure}[t]
\centering
		\includegraphics[width=0.8\textwidth]{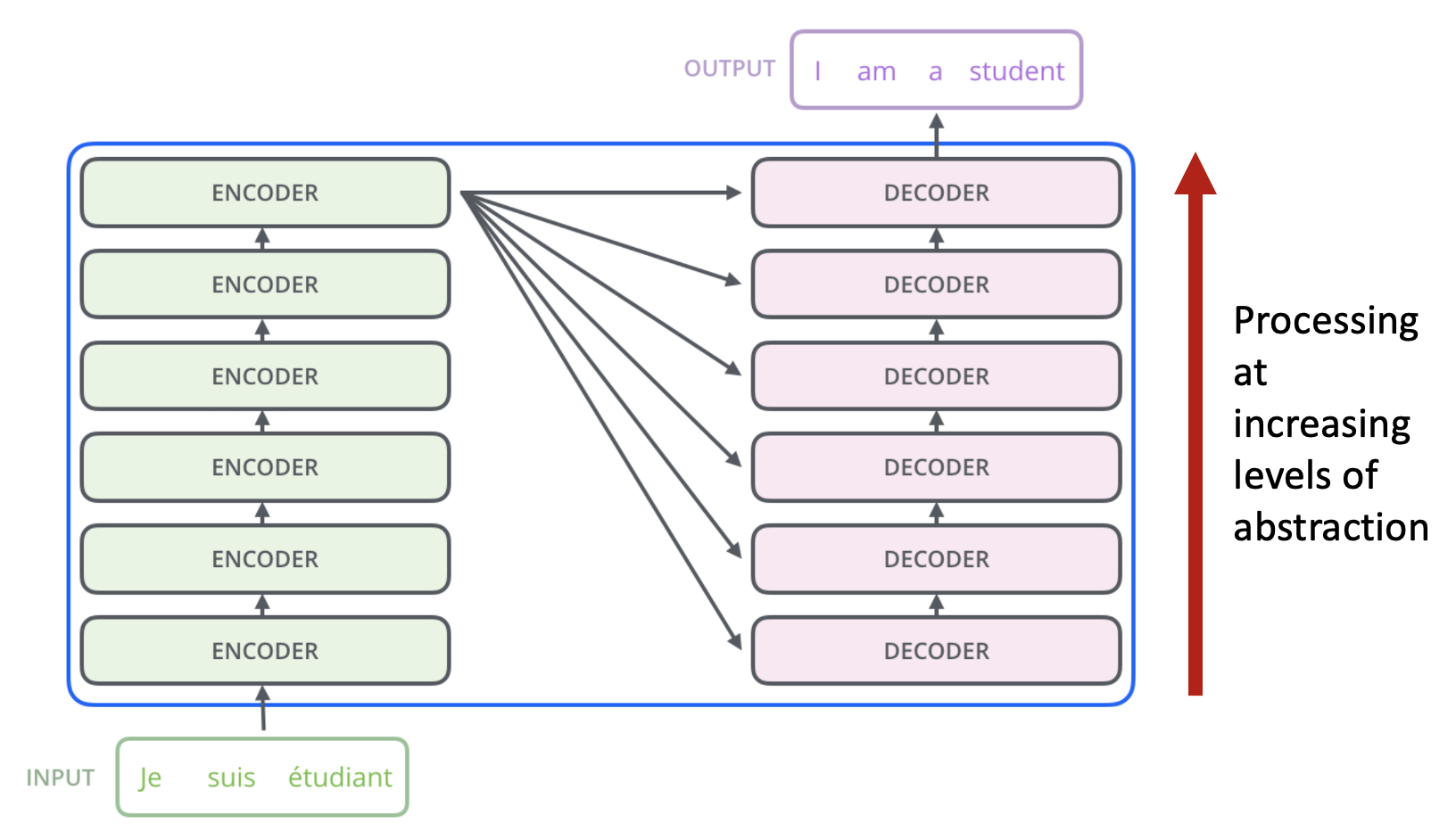}
		\caption{As we proceed up the decoder architectures of a Transformer, the embeddings are increasingly at a higher level from the input embeddings. I posit that this means that they will be processed at more and more abstract levels. For instance, if the initial input embeddings are just purely word level, the middle level embeddings would have taken into account neighbouring words and be more at the sentence level, while those at the top level would be more abstract and related to the concepts expressed by the text rather than the text itself. Image taken taken from: \url{https://jalammar.github.io/illustrated-transformer/} \citep{jalammar-transformer}}
		\label{fig:Transformer}
\end{figure}

\textbf{Iterative processing of semantic meaning.} Moreover, the Transformer architecture actually allows the embeddings of each token to be infleunced by the most similar and closest neighbours via self-attention (via a combination of token embeddings plus position embeddings), which allows for the input representation to be refined in an iterative fashion, solving the case of ambiguous inputs or polysemy (multiple meanings of the same word). Such a hierarchical structure is illustrated in Fig. \ref{fig:Transformer}. 

\textbf{Feedback connections.} I always believe that current deep learning methods suffer from lack of feedback connections to ground the lower levels of processing - it is widely known that in the brain neurons do not just exist in feedforward connections but also have a lot of feedback connections as well. However, recently, observing that increasing the size of Transformers was already sufficient to achieve better and better performance, such as in GPT3.5 and GPT4, I start to wonder if there is indeed a way for Transformers to ground the earlier layers' processing in the later layers' processing. I hypothesise that it is actually able to do some form of feedback grounding, because of the skip connections present between decoder blocks, as illustrated in Fig. \ref{fig:Transformer-GPT}. The embeddings at the lower levels can actually be passed all the way to the later layers (largely unchanged except for LayerNormalisation, which affects all embeddings similarly), and can be processed in the same layer with potential grounding by the embeddings of the later layers. This is extremely powerful, and can actually ground the input processing with knowledge gained at the later part. For instance, in the text "The following did not happen: John went to the market and bought a bunch of eggs, vegetables and meat.", we are able to interpret the entire text in the opposite semantic meaning just because of the words "The following did not happen" at the beginning of the sentence. In fact, as will be discussed in the next section, this presence of skip connections may be the way prompting and grounding in earlier context is so effective in LLMs.

\begin{figure}[t]
\centering
		\includegraphics[width=0.8\textwidth]{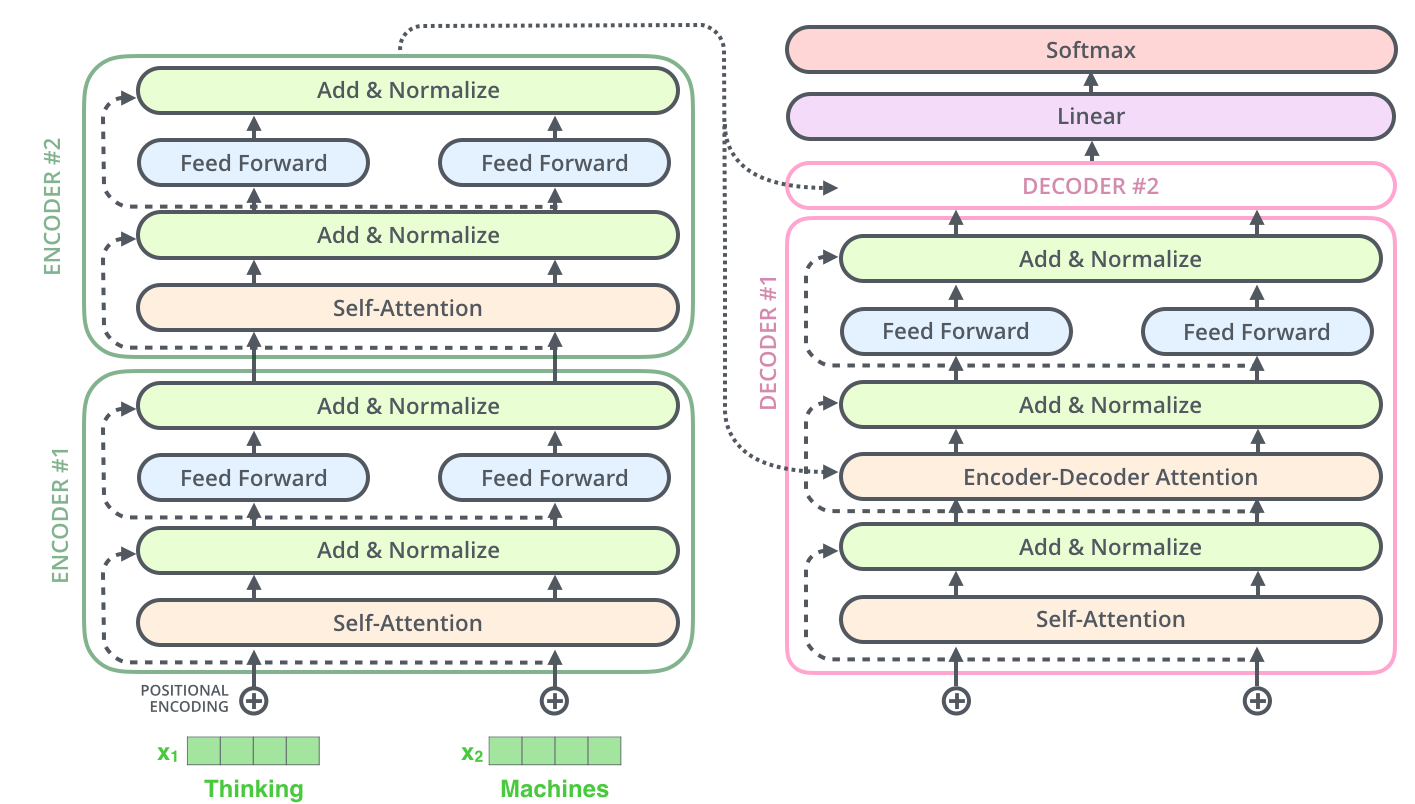}
		\caption{There are skip connections in the encoder and decoder blocks of the Transformer. I posit that this helps the earlier input embeddings to be influenced by later layers in the Transformer, as such there can be some grounding of the later layers embeddings in the earlier layers' processing. Image taken taken from: \url{https://jalammar.github.io/illustrated-transformer/} \citep{jalammar-transformer}}
		\label{fig:Transformer-skip connections}
\end{figure}

\section{Prompting and Zero-shot/Few-shot learning}

Given that LLMs are seemingly able to perform inference at multiple scales of abstraction (see earlier section), this opens an avenue of approaches whereby we can just tell the LLM what we want to do in natural language, and use it to ground the generation. Such an instruction-based method of conditioning generations has proven useful in multiple natural language tasks, as shown in the usage of LLMs flexibly by just an instruction to prompt the task in GLUE \citep{wang2018glue} and SuperGLUE \citep{wang2019superglue} benchmarks.

\subsection{Zero-shot learning}
LLMs are also able to do zero-shot learning very well. For instance, it is able to do zero-shot classification of new contexts simply by using semantic meaning of the tokens it has encountered during training: \\
\framebox{\parbox{\dimexpr\linewidth-2\fboxsep-2\fboxrule}{\itshape%
"You are a classification model meant to classify the context of an input. \\

Context A: In the garden \\
Context B: In the hospital \\
Context C: In the mountains \\
Context D: In the sky \\

Give the contexts for the following inputs: \\
1. Wow, the clouds are so fluffy today \\
2. The IV drip is running out, get a nurse \\
3. The sheep on the pasture are so pretty \\
4. Have you watered the flowers today? \\

Return in the following form: \\
{Number}: {Context Letter}"
}} 

ChatGPT (GPT3.5, May 3 2023 version) returns the following output, which are in general correct:\\
\framebox{\parbox{\dimexpr\linewidth-2\fboxsep-2\fboxrule}{\itshape%
1. D: In the sky \\
2. B: In the hospital \\
3. C: In the mnountains \\
4. A: In the garden
}}

\newpage
\subsection{Few-shot learning}
LLMs are also able to do few-shot learning pretty reliably. For instance, it is able to do few-shot classification of odd and even numbers from just a few sample input and output pairs. In order for to generate consistently, it needs to be given the framework of what the task is about and the possible outputs to ground the generation. Here is the example prompt given: \\

\framebox{\parbox{\dimexpr\linewidth-2\fboxsep-2\fboxrule}{\itshape%
You are a classification machine meant to classify between output A and B. \\

Input: 5 \\
Output: A \\

Input: 7 \\
Output: A \\

Input: 8 \\
Output: B \\

Input: 10 \\
Output: B \\

Input: 13 \\
Output:
}}

ChatGPT (GPT3.5, May 3 2023 version) returns the following output, which is correct:
\framebox{\parbox{\dimexpr\linewidth-2\fboxsep-2\fboxrule}{\itshape%
A
}}

Hence, a trained LLM has shown that it can be equipped with the knowledge of a new task either through zero-shot description-based prompting, or few-shot example-based prompting, and can be the basis of a fast learning system that is adaptive to real-world inputs. Given the quick learning ability of LLMs via prompting, it is no wonder why prompt engineering quickly became very popular following the rise of larger LLMs.

\section{Getting the LLM to reverse engineer the instruction}
LLMs are actually capable of observing multiple input-output pairs and coming up with an instruction to derive the relation between them \citep{zhou2022large}. Furthermore, the Language annotated Abstraction and Reasoning Corpus (LARC) showed that 88\% of the original ARC tasks can be represented in a text instruction where another human can solve it without needing the input-output examples \citep{acquaviva2021larc}. Another paper has also highlighted the efficiency of prompt-based instructions, as one prompt can be worth 100s of training examples on various classification tasks \citep{scao2021many}.

The difficulty of the ARC challenge is that the machine (or human) needs to infer instructions based on limited examples. These instructions are usually difficult to deduce, as one needs to find the pattern with very few sample input-output pairs. However, once the instruction is deduced, it is very easily communicable to other humans using text. Hence, we reframe the ARC Challenge with the following steps:

\begin{enumerate}
    \item Deduce the input-output mapping rule using the LLM from the input-output examples
    \item Apply this rule to the test input to get the test output
\end{enumerate}

\section{Chain of Thought}

It is often difficult to do planning on complicated tasks which involve multiple steps. The ARC Challenge sometimes also involves multiple manipulations of the input image in order to derive the output. For this kind of problems, we can utilize approaches such as Chain of Thought (CoT) prompting \citep{wei2022chain}, which uses demonstrations of details like the steps for mathematical computation to train the language model. Moreover, we do not even need to provide the human-labelled detailed demonstration as shown in the CoT paper, but can get the LLM to generate its own thoughts. The "ReAct: Synergizing reasoning and acting in language models" paper shows how one way of prompting the LLM for it to generate detailed thoughts and act upon it \citep{yao2022react} - using the Thought, Action, Observation framework. 

\textbf{Hierarchical Planning.} CoT is still a largely linear way to do planning, as it involves having the previous action or plan before generating the next one. More recently, LLMs have been utilised in a hierarchical fashion, whereby the first step involves coming up with the broad plan, and the second step is to come up with the details. This is utilised in HuggingGPT \citep{shen2023hugginggpt} and AutoGPT \citep{autogpt} to generate an overall plan before breaking down into the detailed steps. This way of hierarchical planning was also used in the Generative Agents paper \citep{park2023generative} to generate a detailed action plan for an agent's day. 

This approach of hierarchical planning is actually quite similar to how humans think. We do not have a detailed plan of our day right at the beginning, but think in a broad way like doing work in the morning, lunch, meet friends in afternoon, home in the evening and so on. Then, when prompted \textbf{why} do you want to do this, we go up a layer of abstraction to think about the goals of our lives. When prompted \textbf{how} do you want to do this, we go down a layer of abstraction to think about the specifics of the various plans of our lives. Hence, explicitly prompting the LLM to come up with the broad plan, and then using the broad plan to ground the generation for the detailed plan is a promising approach. It also helps circumvents the problem of the LLM having limited planning abilities, as we can plan the broad steps first, which are usually much shorter than the entire sequence of detailed steps.

\section {Grounding in Human Biases}

The ARC Challenge is difficult for computers because there is a huge number of possibilities to interpret high-dimensional real-world data, but easy for humans because humans can curate the possibilities based on some innate biases, like that of the Gestalt principles \citep{todorovic2008gestalt}. In fact, without such innate biases, it can be difficult for anyone to learn quickly in the real world. \cite{vallortigara2021born} wrote a book, "Born Knowing", which highlights that chicks come born with plenty of innate biases like preference for animate objects, which could help them learn faster. Similarly, human newborns come with a preference for face-like objects to help with recognition of the mother. Some human behaviours like suckling are also innate, rather than learnt, to facilitate survival.

Alas, we may not be born \textit{tabular rasa} like what is done in AlphaZero \citep{silver2018general}. In experiments with AlphaZero, it takes weeks with a single GPU just to learn how to play well enough to win a human \citep{min2022brick} in a 4-in-a-row Tic-Tac-Toe game in a 7x7 grid with an unplayable position. Simply changing the unplayable position was enough to cause AlphaZero to become weaker than humans, and extensive training of various random unplayable positions was required for it to learn. Hence, for generalisability, pursuing optimality in Reinforcement Learning from a clean slate like that in static games like Chess or Go may not be the way to go. Rather, we need to ground the possibilities of what we need to do or interpret perception with some innate bias or some past experience in order to learn fast and be generalisable.

Since LLMs like GPT4 are not able to be trained to a new set of input-output due to constrains of API, we utilise prompting to instill the human biases required for the machine to reduce the possibilities of interpreting the input-output pairs of the ARC Challenge.

\newpage
\section{Naïve Method (Single Prompt)}
Given that LLMs have proven effective at learning an arbitrary task just by prompting, we try to do a naïve method of getting it to solve ARC tasks just from a single prompt alone. This prompt should be as generalisable as possible and should not be fine-tuned to any single one task.

Using the above ideas of grounding in human biases, CoT prompting and getting LLMs to come up with broad descriptions, then detailed steps, and then using the detailed steps to map from test input to test output, we come up with an example prompt for ARC as given below:

\framebox{\parbox{\dimexpr\linewidth-2\fboxsep-2\fboxrule}{\itshape%
“You are given a series of inputs and output pairs. \\
These are all in the form of a 2D array, representing a 2D grid, with values from 0-9. \\
The values are not representative of any ordinal ranking. \\
Input/output pairs may not reflect all possibilities, you are to infer the simplest possible relation making use of symmetry and invariance as much as possible. \\

The input can be something like: \\
> entire grid being the sandbox to manipulate \\
> using a part of the grid (individual squares or portions of the grid) to depict instructions of how to do the task. symmetry is important. \\
> using regions of similar value to depict area for answer of the task \\

The output can be something like: \\
> same output size as input after performing action \\
> output one of the fixed predetermined patterns used to classify the input image \\
> using output to show the ordering of objects, such as by size, height, width, position, value \\

Each of the input-output relation can be done with one or more actions chained together, which could be something like (not exhaustive): \\
$-$ object view (defined as continuous squares connected horizontally, vertically and/or diagonally, separated by 0 values) \\
> objects can be of the same value, or different values combined together \\
> objects may be hidden beneath other objects \\
> rotating or shifting objects \\
> changing value of object \\
> objects can be manipulated and mapped to a different number of output squares \\
> different objects may be manipulated differently based on context \\

$-$ overall view \\
> rotation / reflection symmetry \\
> continuation of a pattern \\
> changing values \\

$-$ segment view \\
> combine two segments of the input into one single one based on a simple rule \\
> rule can be certain values are prioritized over others, or combination of values into new ones \\

Do the following: \\
$-$ What is the broad description of the input/output relation that holds for all input/output pairs? \\
$-$ What is the step by step description of the input/output relation that holds for all input/output pairs? \\
$-$ Apply this description to the test input and find out the answer 'to\_be\_filled'.”\\\\ 

[Insert .json for task here with all the input-output pairs in json format, with the test output replaced by 'to\_be\_filled']
}}

The method to derive the json format is simply replacing the output section of the original json format from the ARC Challenge 2 dataset with 'to\_be\_filled'. The code to do so can be found here: \url{https://github.com/tanchongmin/ARC-Challenge/blob/main/arc_challenge.ipynb}

\subsection{Example: Public Evaluation Task 157 (66e6c45b.json)}

Here is the modified .json for this task without test output: \\
\framebox{\parbox{\dimexpr\linewidth-2\fboxsep-2\fboxrule}{\itshape%
\{'train': [\{'input': [[0, 0, 0, 0], [0, 3, 4, 0], [0, 7, 6, 0], [0, 0, 0, 0]], 'output': [[3, 0, 0, 4], [0, 0, 0, 0], [0, 0, 0, 0], [7, 0, 0, 6]]\}, \{'input': [[0, 0, 0, 0], [0, 5, 6, 0], [0, 8, 3, 0], [0, 0, 0, 0]], 'output': [[5, 0, 0, 6], [0, 0, 0, 0], [0, 0, 0, 0], [8, 0, 0, 3]]\}], 'test': \{'input': [[0, 0, 0, 0], [0, 2, 3, 0], [0, 4, 9, 0], [0, 0, 0, 0]], 'output': 'to\_be\_filled'\}\}
}}

Here is the output for Public Evaluation Task 157 (66e6c45b.json) using GPT3.5:\\
\framebox{\parbox{\dimexpr\linewidth-2\fboxsep-2\fboxrule}{\itshape%
Broad Description:
The input/output relation involves manipulating the input grid by applying a series of transformations based on certain rules and symmetries to produce the output grid.

Step-by-Step Description:

Identify any objects in the input grid based on contiguous squares with the same value.
Apply transformations to the identified objects based on their position, symmetry, and value. These transformations may include rotating, reflecting, shifting, or changing the value of the object.
Combine or split objects as needed based on certain rules or conditions.
Repeat steps 1-3 until the output grid is produced.
For this specific test input, we can apply the following transformations to obtain the output grid:

Identify two objects with values 2 and 3 respectively.
Rotate the object with value 2 by 90 degrees counterclockwise and shift it to the right to fill the empty space.
Rotate the object with value 3 by 90 degrees clockwise and shift it to the left to fill the empty space.
Combine the two objects to form the output grid:
[[2, 0, 0, 3],
[0, 0, 0, 0],
[0, 0, 0, 0],
[4, 0, 0, 9]]
Therefore, the answer is [[2, 0, 0, 3], [0, 0, 0, 0], [0, 0, 0, 0], [4, 0, 0, 9]].
}}

\begin{figure}[h]
\centering
		\includegraphics[width=0.5\textwidth]{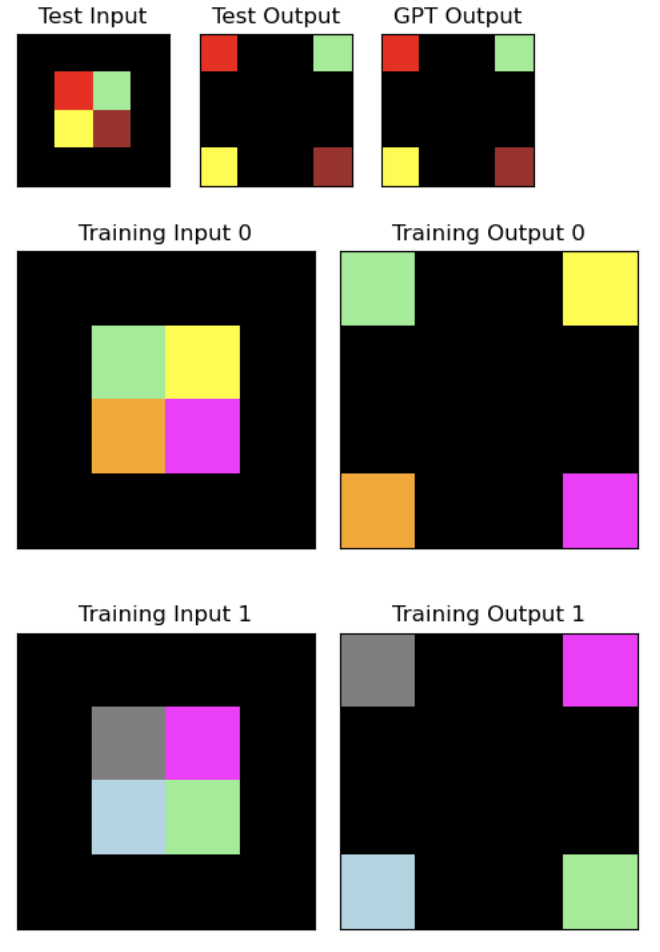}
		\caption{The Test Output and GPT Output for Public Evaluation Task 157 (66e6c45b.json)}
		\label{fig:157}
\end{figure}

\newpage
\subsection{Example: Public Evaluation Task 162 (68b67ca3.json)}

Here is the modified .json for this task without test output: \\
\framebox{\parbox{\dimexpr\linewidth-2\fboxsep-2\fboxrule}{\itshape%
\{'train': [\{'input': [[2, 0, 2, 0, 1, 0], [0, 0, 0, 0, 0, 0], [2, 0, 1, 0, 0, 0], [0, 0, 0, 0, 0, 0], [3, 0, 0, 0, 0, 0], [0, 0, 0, 0, 0, 0]], 'output': [[2, 2, 1], [2, 1, 0], [3, 0, 0]]\}, \{'input': [[3, 0, 0, 0, 2, 0], [0, 0, 0, 0, 0, 0], [8, 0, 0, 0, 8, 0], [0, 0, 0, 0, 0, 0], [0, 0, 1, 0, 0, 0], [0, 0, 0, 0, 0, 0]], 'output': [[3, 0, 2], [8, 0, 8], [0, 1, 0]]\}, \{'input': [[1, 0, 0, 0, 0, 0], [0, 0, 0, 0, 0, 0], [0, 0, 2, 0, 0, 0], [0, 0, 0, 0, 0, 0], [6, 0, 0, 0, 6, 0], [0, 0, 0, 0, 0, 0]], 'output': [[1, 0, 0], [0, 2, 0], [6, 0, 6]]\}], 'test': \{'input': [[3, 0, 3, 0, 4, 0], [0, 0, 0, 0, 0, 0], [7, 0, 0, 0, 1, 0], [0, 0, 0, 0, 0, 0], [7, 0, 0, 0, 1, 0], [0, 0, 0, 0, 0, 0]], 'output': 'to\_be\_filled'\}\}
}}

Here is the output for Public Evaluation Task 162 (68b67ca3.json) using GPT4:\\
\framebox{\parbox{\dimexpr\linewidth-2\fboxsep-2\fboxrule}{\itshape%
Broad description of the input/output relation:

The input/output relation involves extracting non-zero elements from specific positions in the input grid and arranging them into a smaller 3x3 output grid, maintaining their relative positions.

Step by step description of the input/output relation:

Identify non-zero elements in the following positions of the input grid: (1,1), (1,3), (1,5), (3,1), (3,3), (3,5), (5,1), (5,3), and (5,5).
Place these non-zero elements into the corresponding positions in a 3x3 output grid.
Applying this description to the test input:

Test input:
[[3, 0, 3, 0, 4, 0],
[0, 0, 0, 0, 0, 0],
[7, 0, 0, 0, 1, 0],
[0, 0, 0, 0, 0, 0],
[7, 0, 0, 0, 1, 0],
[0, 0, 0, 0, 0, 0]]

Following the step-by-step description, we extract the non-zero elements from the specified positions:
(1,1): 3, (1,3): 3, (1,5): 4, (3,1): 7, (3,3): 0, (3,5): 1, (5,1): 7, (5,3): 0, (5,5): 1

Place these elements into the 3x3 output grid:
[[3, 3, 4],
[7, 0, 1],
[7, 0, 1]]

So, the output for the test input is:
[[3, 3, 4],
[7, 0, 1],
[7, 0, 1]]
}}

\begin{figure}[h]
\centering
		\includegraphics[width=0.3\textwidth]{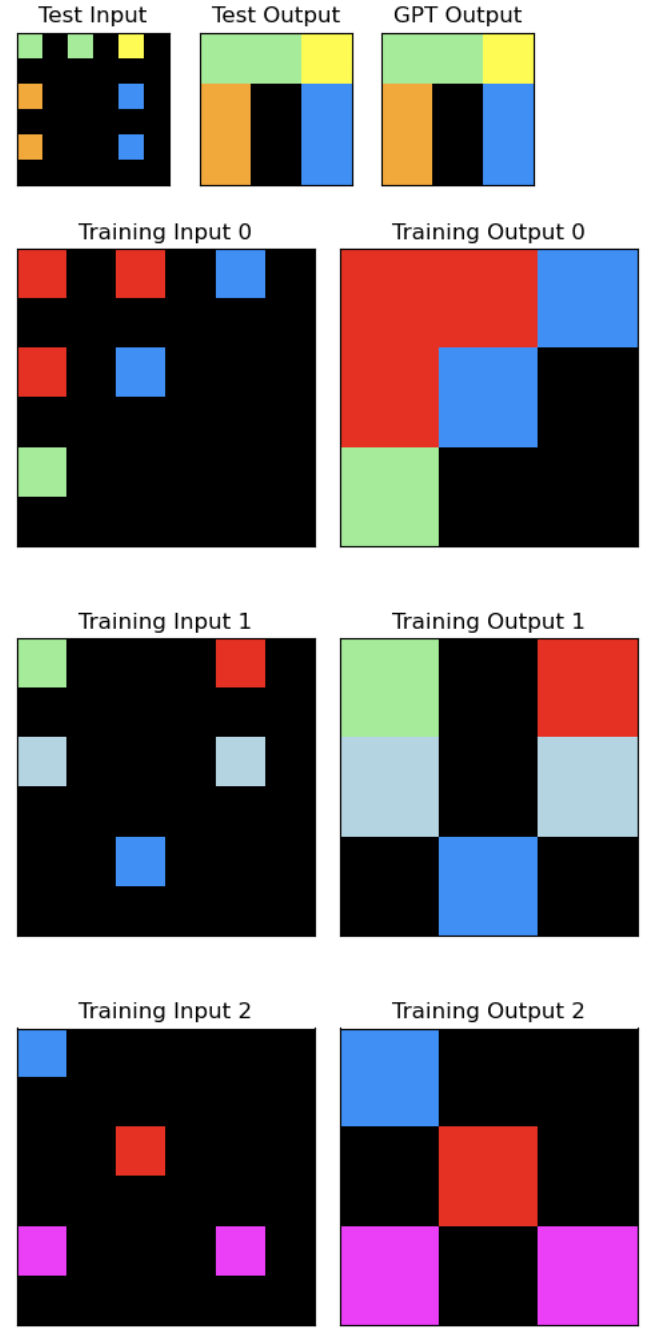}
		\caption{The Test Output and GPT Output for Public Evaluation Task 162 (68b67ca3.json)}
		\label{fig:162}
\end{figure}

\newpage
\subsection{Evaluation}
This naïve approach has some success with the smaller ARC tasks. So far, with limited testing, this naïve method on GPT3.5 or GPT4 has solved the following tasks out of 4 tested tasks on the Evaluation set: 157 (66e6c45b.json), 162 (68b67ca3.json). These two tasks have failed, although only slightly and are likely to be solved with more specific prompt engineering: 158 (66f2d22f.json), 170 (6ea4a07e.json). See the testing of GPT4 on the ARC Challenge via this url: \url{https://www.youtube.com/watch?v=vt2yG1da8Fg}. With some more fine-tuning of the actions that can be performed, I believe we can get it to work for more tasks. The key takeaway is that prompting can help to ground the model to think of feasible solutions it would otherwise not have. 

That said, the json input for the 2D array is not a great one to extract object-level relations, and the prompt needs to continuously ask for GPT4 to think of the input as an object. The prompt is intended to be very generic and gives the broad input-output relation, along with some tips as to how prior ARC puzzles can be solved. As GPT4 is not that great at doing detailed planning, we follow the hierarchical approach done by HuggingGPT \citep{shen2023hugginggpt} or AutoGPT \citep{autogpt}, and ask the model to list out the broad description first. Thereafter, after being grounded by the broad description, the model then generates the detailed step by step description. This description is then used to get the answer by applying these steps to the test input.

Initially, I tried to get GPT4 to output a Python program to handle the manipulation from input to output. While this could work for simple problems, in general, I find that the program output generated may be different from the intention in the step by step description, and in general, the step by step description in words was more accurate. As such, the example prompt above did not ask for a program output from GPT4.

\section{Improvements to the Naïve Method}

Following my experiments with the naïve method, I have identified the following issues:

\begin{enumerate}

\item Limited understanding of what an object is from the json file

\item Limited context length to store json representations of large grids / multiple input-output samples

\item Limited context length to store instructions

\item Limited fact-checking abilities to determine if the input-output relation derived is correct

\end{enumerate}

\begin{figure}[h]
\centering
		\includegraphics[width=0.8\textwidth]{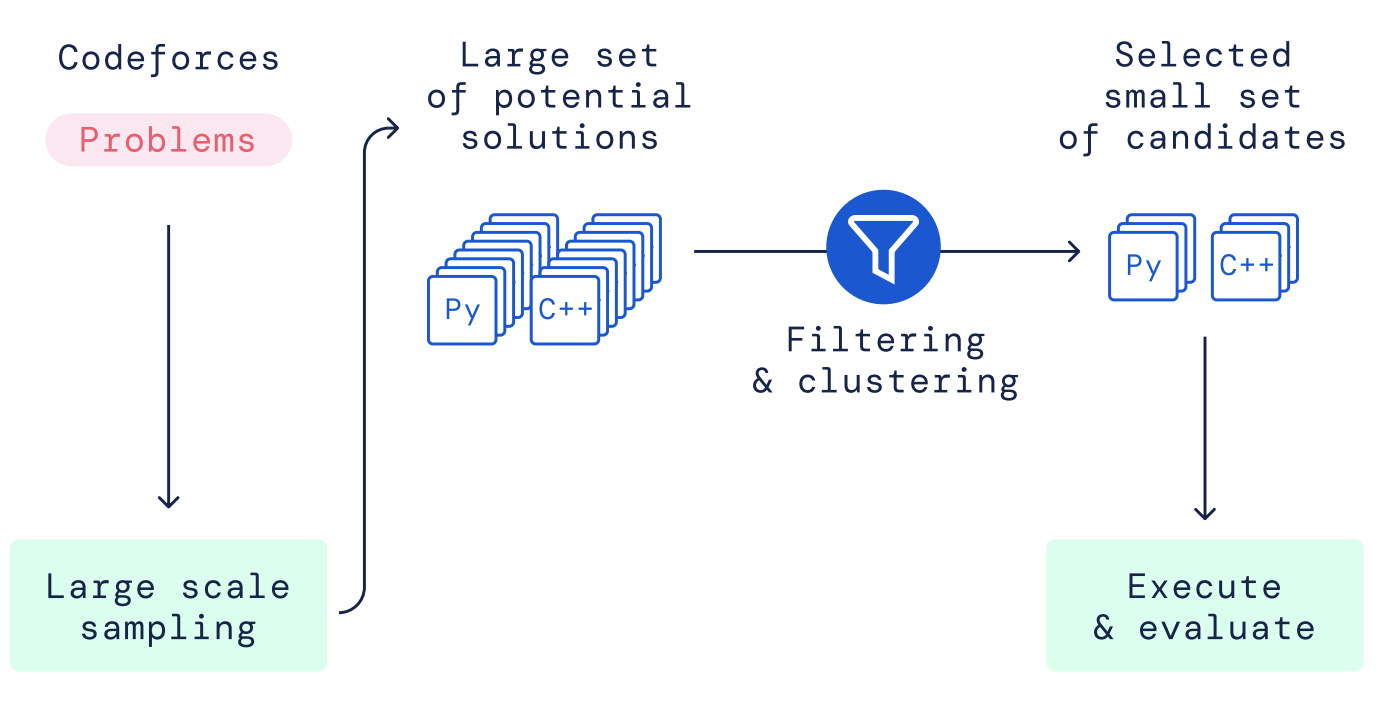}
		\caption{Large-scale sampling and filtering in AlphaCode. Extracted from Fig. 4 in \cite{li2022competition}}
		\label{fig:alphacode}
\end{figure}

These are the potential solutions to the above issues:

\begin{enumerate}
    \item In order to do the ARC challenge well, it would be good to imbue in the model a sense of what an object is, and also how images look like in the real world. This is because there are some ARC challenges which use concepts like object permanence, gravity, which would be present in real world situations but not for a computer which is only trained on pixels in the ARC challenge. As such, we could take a leaf from the Visual Question Answering (QA) domain \citep{wu2017visual}, and give the LLM the ability to ask questions about the input and output images and iteratively refine its input-output relation based on it. This Visual QA could be done with the base model as images in the wild, but should be fine-tuned with past ARC challenge data, as the distribution of pixel information between the real world and ARC challenge dataset may be different, though the concepts may be the same. My hypothesis is that pixel-based representation may be too high-dimensional to model the world, hence, being able to compress it down to low-dimensional text via Visual QA would be a huge plus for interpretability. 
    \item Instead of putting all the input-output examples in the same json, we can separately ask the model to give a description for each of the input-output pairs. Then, we can prompt another model to find similarities between each of the descriptions of these input-output pairs and collate to a general input-output representation.
    \item Instead of having only one GPT model to give the prompt for the instructions, we could split the prompt into multiple parts. For instance, the object view can be one model, the overall view can be another model, the segment view can be another model, and so on. This would mean that we can ground the instructions in more fine-grained action space that would increase the likelihood of solving the ARC challenges. We can then select the best performing instruction, by asking the various models to come up with different sets of input-output instructions, and collating them into one pool of potential instructions. Then, we can evaluate all of them and use the best one.
    \item We could have a separate GPT model to evaluate the input-output mapping. This model takes in the pool of potential instructions generated by the above steps, and evaluates them one by one. The moment any of the instructions fails to generate the input-output map of the training cases, it is discarded. This approach of generating more potential mappings and discarding them based on grounding by the training set is used in AlphaCode, where they generate multiple programs by simply changing the hyperparameters or by using more random generations of the LLM, and then eliminate those non-performant ones that do not give the right output in the training cases \citep{li2022competition} (See Fig. \ref{fig:alphacode} for an illustration). Currently, I envision this model to take in just the instruction and the input json, and output the json after the instruction and check that it matches with the actual output json. An alternative is to ask GPT4 to come up with the Python code to do the input-output mapping, and then run the code to check for correct output - I suspect this may be inferior due to problems mapping to the right Python program from the instruction.
    
\end{enumerate}

\section {GPT as a System}

With the recent trends of utilising multiple LLMs together as a system, such as in AutoGPT \citep{autogpt}, it could potentially allow the model to scale better by off-loading various tasks to different LLM models, and letting all these models work together in a large ecosystem. Such a model is outlined in the Improvements section above, and more can be tuned in order to make the system as performant as possible.

\section {Memory as the way ahead}

Given that we are not able to train the weights of GPT to fit to the training set of the ARC challenge, using memory is the best way to go about imbuing the model with learnt knowledge. Humans learn very fast because we have memory to ground our current experiences and we can choose the best action based on what we have seen in the past. For example, if I see a snake on Path A, I will avoid Path A next time and choose Path B instead. This instantaneous way of learning is something that is not natural in deep learning, as it typically takes hundreds or thousands or more iterations in order to update the weights sufficiently so that it can learn well for Deep Learning, such as in Deep Reinforcement Learning. A more detailed explanation can be found in "Learning, Fast and Slow" \citep{john2023learning}. 

Currently the naïve method does not use memory of what has been stored earlier. If we were to use memory, I posit the best way to use it will be via text descriptions of the broad and detailed input-output relations stored from the earlier training examples. This would make the memory more generic rather than storing memory of the images. We then have two memories of instructions, one I call \textit{BroadInstruct}, and the other \textit{DetailedInstruct}, which details the broad description and detailed steps of earlier instructions from earlier ARC tasks. I can envision a system using it to be as such:

\begin{enumerate}
    \item Use the naïve method to determine the broad description of the task
    \item From the broad description, retrieve from a database (e.g. Pinecone) using OpenAI Vector Embeddings \citep{neelakantan2022text} or similar embeddings to retrieve the top $k$ neighbours from \textit{BroadInstruct}. $k$ is a hyperparameter that can be tuned, and can be set to 5 by default.
    \item Conditioned on the top $k$ neighbours as context, perform retrieval-augmented generation \citep{lewis2020retrieval} to generate the refined broad description of the task
    \item Repeat the earlier steps until convergence
\end{enumerate}

Now, having generated the broad description of the task, we move on to generate the detailed steps.
\begin{enumerate}
    \item Use the naïve method with the broad descrption as context to determine the detailed steps of the task
    \item From the generated detailed steps, retrieve from a database (e.g. Pinecone) using OpenAI Vector Embeddings \citep{neelakantan2022text} or similar embeddings to retrieve the top $k$ neighbours from \textit{DetailedInstruct}. $k$ is a hyperparameter that can be tuned, and can be set to 5 by default.
    \item Conditioned on the top $k$ neighbours as context, perform retrieval-augmented generation \citep{lewis2020retrieval} to generate the refined detailed steps of the task
    \item Repeat the earlier steps until convergence
\end{enumerate}

Hence, we can utilise past knowledge of earlier ARC tasks for more accurate conditioning of the broad description and detailed steps needed for future ARC tasks. If the task is solved, we can then add in this broad and detailed description into \textit{BroadInstruct} and \textit{DetailedInstruct} respectively.

Apart from imbuing learning ability, retrieval-augmented generation has an added benefit of increasing the consistency of the LLM-generated output, as it is more in line with what is required, which may be helpful with getting the right solution in fewer generations. For more complicated problems (more complex than ARC challenge), in order to constrain memory storage given a limited storage space, we can also selectively store memory based on how surprising it is and how "emotional" the experience is. These can be explored in future challenges where there is too much perceptual information and memory storage is a constraint, but for ARC, I believe we can just keep all the memory as the number of ARC tasks are not large.

\section{Conclusion}

Overall, the ARC challenge is a very unique one, and can serve to pave the way for systems that are fast learning and can generalise well to arbitrary tasks. With the right innate biases via prompting, the right hierarchical structure to condition generation of detailed steps from broad description, a multi-agent architecture to split long prompts up into performant smaller sub-systems, a better way to interpret images using Visual QA, as well as better learning and grounding in past memory, I posit that GPT4 can eventually be made to solve the majority of the ARC tasks.

\bibliography{neurips}
\bibliographystyle{neurips}

\end{document}